\documentclass{article}

 \usepackage[nonatbib,preprint]{neurips_2019}

\usepackage[utf8]{inputenc} 
\usepackage[T1]{fontenc}    
\usepackage{hyperref}       
\usepackage{url}            
\usepackage{booktabs}       
\usepackage{amsfonts}       
\usepackage{nicefrac}       
\usepackage{microtype}      

\usepackage{booktabs} 
\usepackage{subfig}
\usepackage{amsmath}
\usepackage{mathtools}
\usepackage[toc,page]{appendix}
\usepackage[utf8]{inputenc}
\usepackage[T1]{fontenc}
\usepackage{lmodern}
\usepackage{enumitem}
\usepackage{algorithm}
\usepackage[noend]{algpseudocode}

\title{A Hybrid Algorithm for Metaheuristic Optimization}

\author{
  Sujit Pramod Khanna \thanks{\url{https://people.rit.edu/xxxx/home.html}} \\
  \texttt{sk2913@rit.edu} \\
  \And
  Alexander Ororbia II\thanks{\url{https://www.cs.rit.edu/\~ago/}} \\
  \texttt{ago@cs.rit.edu} \\
  \and
  Golisano College of Computing and Information Sciences\\
  Rochester Institute of Technology\\
  Rochester, NY 14623 \\
}

\begin{document}

\maketitle

\begin{abstract}
\label{sec:abstract}
We propose a novel, flexible algorithm for combining together metaheuristic optimizers for non-convex optimization problems. Our approach treats the constituent optimizers as a team of complex agents that communicate information amongst each other at various intervals during the simulation process. The information produced by each individual agent can be combined in various ways via higher-level operators.  In our experiments on key benchmark functions, we investigate how the performance of our algorithm varies with respect to several of its key modifiable properties. Finally, we apply our proposed algorithm to classification problems involving the optimization of support-vector machine classifiers. \end{abstract}	


\section{Introduction}
\label{sec:intro}

The class of nature-inspired metaheuristic optimization algorithms generally do not make use of gradients of an objective function when searching for optima. These algorithms are generally population-based, where agents within the population improve their fitness iteratively, ultimately contributing to the collective's overall generalization performance. However, while each metaheuristic overcomes various limitations in the optimization process with respect to certain types of data spaces, the No Free Lunch Theorem states that no single one of these algorithms will perform best across all problems. This motivates the use of ensembling, where we might exploit different favorable properties and generalization abilities of each algorithm collectively, ultimately improving robustness as we are able to handle a wider variety of situations. However, work in ensembling in the space of metaheuristic optimization has been rather limited, treating the process as a rather simple, post-processing step, where the final results obtained by the metaheuristics, each working in isolation, are simply aggregated through majority-voting or averaging. 

In this work, we propose a novel, flexible meta-algorithm for ensembling metaheuristic optimizers -- one in which treats the constituent optimizers as a team of complex agents that must interact with one another at various intervals during the simulation process, sharing information, such as their own individual global best-fit candidate solutions, that can be combined in various ways via higher-level operators.  We investigate how the performance of our algorithm varies with respect to several of its core modifiable properties, and apply our proposed approach to classification problems entailing the optimization of parameters of Support Vector Machines. 

\section{Related Work}
\label{sec:rw}
In statistical machine learning, ensemble methods use multiple learning algorithms to obtain better predictive performance than what could be obtained by any single one classifier. The simplest approaches to ensemble learning include bagging \cite{Breiman1996}, boosting \cite{breiman1998}, or Bayesian averaging \cite{Domingos:2000:BAC:645529.657966}. The ``Bucket of Models'' is another approach\cite{Dzeroski04iscombining} which involves a procedure that chooses model/classifier is best for the specific problem at hand. These approaches generally are constructed for statistical classifiers and, very rarely if at all, do not focus on optimization procedures and metaheuristics. Moreover, in most ensembling setups, every classifier is treated independently with no notion of cooperation between different components of the ensemble. In metaheuristic optimization, we argue that a communication scheme that coordinates the various individual procedures, instead of treating them as a simple combination of isolated solvers, might improve our ability to solve complex nonlinear black-box optimization problems.

One key source of inspiration for our aggregation approach comes from multi-agent learning systems, where multiple agents co-operate with each other to solve a specific task \cite{Tan93multi-agentreinforcement}.  Ideas from multi-agent system theory have already proven to be quite powerful when applied to other domains, such as in reinforcement learning \cite{Gupta2017CooperativeMC}, where an indirect cooperative mechanism was proposed for multiple neural agents to communicate by sharing their policy parameters. Related to this was an effort to train cooperative agents in a (deep) reinforcement learning framework utilizing an ensemble of different sub-policies \cite{Lowe2017MultiAgentAF}, which made the training process more robust. In this paper, we treat our proposed algorithm as a multi-agent system as well, where every agent is an individual optimizer and the optimization process is designed in such a way that individual optimization algorithms solve a common problem through collaborative communication.

\section{A Multi-Metaheuristic Optimizer}
\label{sec:algo}
\begin{figure}[t]
\begin{center}
  \includegraphics[scale=0.7]{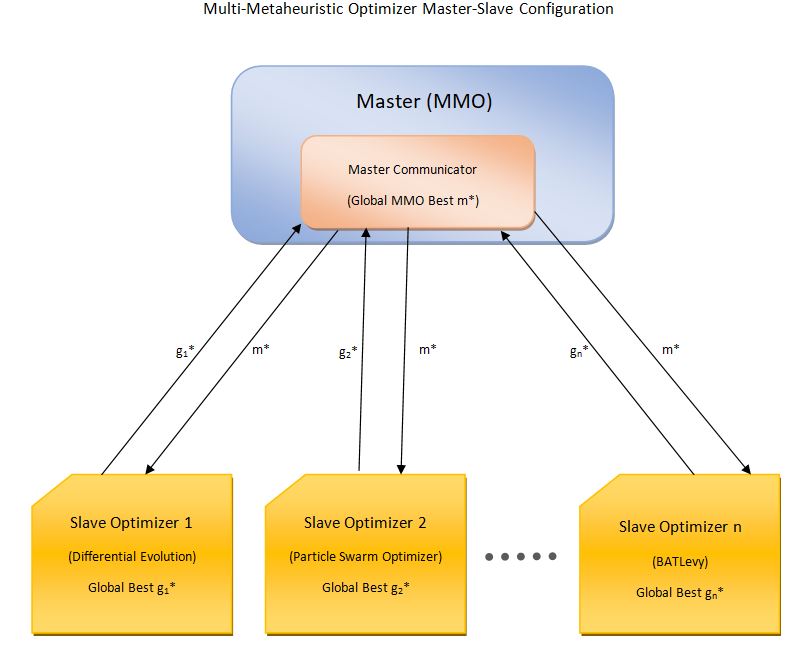}
  \caption{Depiction of the Multi-Metaheuristic Optimizer master-slave configuration.}
  \label{fig:system}
  \end{center}
\end{figure}

In this paper, we propose an aggregation approach to optimization which maintains the consistency and exploits the exploratory power of previously proposed metaheuristics while incorporating communication scheme between them during the optimization process. We call our algorithm, or meta-metaheuristic procedure, the Multi-Metaheuristic Optimizer(MMO). Within MMO, multiple, different optimization algorithms periodically communicate their global best solutions to each other. Such a process can be viewed from a Master/Slave perspective, where the Multi-Metaheuristic Optimizer(MMO) is the master and individual metaheuristics algorithms are the slaves. Figure \ref{fig:system} depicts such a configuration. Furthermore, we have developed a flexible, extensible software package that concretely instantiates our proposed procedure.\footnote{The github repo supporting this work is{MMO} (link will be placed upon publication acceptance).}

\subsection{The Algorithm}
\label{subsec:the algo}

The version of our proposed MMO algorithm we experiment with in this paper contains a set of seven different nature-inspired optimization algorithms, two of which are novel variations we propose based on our prior experimentation with Levy flights. Each algorithm that we implemented within our MMO framework is defined below. The problems these algorithms intend to solve follow the general form: $\mathit{min}( f_1(\mathbf{x},f_2(\mathbf{x}),\cdots,f_j(\mathbf{x})), \mbox{, s.t., } \mathbf{x} \in X$ (in this paper, we set $j=1$, and leave $j > 1$ for future work). $g^*$ is the global best-found solution so far while $x_i^{*(t)}$ is the local best-found solution (for a particular agent $i$).


\begin{itemize}[wide, nosep, labelindent = 0pt, topsep = 1ex]
\item Particle Swarm Optimizer(PSO): PSO \cite{PSO} uses communication between agents (or candidate solution vectors, i.e., ``particles'') by tracking the local best solution each agent has found. Furthermore, PSO tracks a global best solution that has been found so far by the entire population of agents. For each agent, the procedure uses both the global and local best values to calculate a direction/next step. The PSO procedure is mathematically represented as: \newline
\[
\begin{split}
v_i^{t+1} &= v_i^t + \alpha\varepsilon_1[g^* - x_i^t] + \beta\varepsilon_2[x_i^{*(t)} - x_i^t]\\
x_i^{t+1} &= x_i^t + v_i^{t+1}
\end{split}
\]

\item Particle Swarm Optimizer with Lévy Flights (PSOLévy): PSOLévy is similar to PSO, however, we propose an alteration to the global update equation. Instead of using Gaussian noise, we propose injecting noise though a Lévy flight process \cite{Gutowski2001LevyFA}. The resulting algorithm is: \newline
\[
\begin{split}
v_i^{t+1} &= v_i^t + \alpha L(\lambda)[g^* - x_i^t] + \beta\varepsilon_2[x_i^{*(t)}  - x_i^t]\\
x_i^{t+1} &= x_i^t + v_i^{t+1}
\end{split}
\]
where $ L(\lambda)$ is a Lévy process

\item Differential Evolution(DE): DE \cite{DE} is a population-based algorithm and specifically is a vector-operator based variant of a genetic algorithm \cite{holland-genetic-algorithms-1992}. It has the same primary steps as a classical genetic algorithm: mutation, crossover, and selection. These operators are formally represented as: \newline
\[
\begin{split}
v_i &= x_p^t + \alpha(x_q^t - x_r^t)\\
u_{j,i}^{i+1} &=
\begin{cases}
v_{j,i} & r_i \leq C_r, j = 1,\dots,d\\
x_{j,i} & \text{otherwise}
\end{cases}\\
x_i^{t+1} &=
\begin{cases}
u_i^{t+1} & f(u_i^{t+1}) \leq f(x_i^t),\\
x_i^t & \text{otherwise}
\end{cases}
\end{split}
\]
\item BAT Algorithm: The BAT algorithm \cite{BAT} is inspired by echolocation behavior of microbats and usefully formalizes this in a way such that it is associated with the cost/objective function of interest. A BAT algorithm adjusts the amplitude $A_{i}^{t}$ and pulse emission $r_{i}^{t}$ of its individual agents (candidate solution vectors) to hone in on optimal locations. The procedure has two update equations, one for global search and another for local search. These equations are defined as follows: \newline
\[
\begin{split}
    f_i = f_{min} + \beta[f_{max} - f_{min}]\\ x_i^{t+1} = x_i^t + v_i^{t} + f_i[x_i^t - g^*]\\ x_{new} = g^* + \sigma\varepsilon A^{(t)}\\ A_{i}^{t+1}=\alpha A_{i}^{t}, r_{i}^{t+1}=r_{i}^{0}[1 - e^{\gamma t}] 
\end{split}
\]

\item BAT Algorithm with Lévy Flights (BATLévy): BATLévy our proposed modification of the BAT algorithm. Much as in PSOLévy, the local search equation is augmented with a Lévy flight process instead of a Gaussian noise process. Formally, this yields the following: \newline
\[
\begin{split}
    f_i = f_{min} + \beta[f_{max} - f_{min}]\\ x_i^{t+1} = x_i^t + v_i^{t} + f_i[x_i^t - g^*]\\ x_{new} = g^* + \sigma L(\lambda) A^{(t)}\\ A_{i}^{t+1}=\alpha A_{i}^{t}, r_{i}^{t+1}=r_{i}^{0}[1 - e^{\gamma t}] 
\end{split}
\]

\item Cuckoo Search (CS): CS \cite{CS} is a metaheuristic procedure based on the brood parasitism of cuckoo birds. Noting that \(p_a\) is the ratio of solutions that will be removed every iteration, the algorithm works as follows:
\begin{itemize}
    \item A population of nests/solutions is initialized with the equation:
    \[x_i^{t+1} = x_i^t + \alpha s \otimes H(p_a - \varepsilon) \otimes (x_j^t - x_k^t)\]
    \item A random solution(cuckoo egg) is generated using Lévy flights, say \(x_j\).Then \(x_i+1\) and \(x_j\) are compared and the best out of them is kept.
    \[x_{i+1}^t = x_i^t + \alpha L(s, \lambda)\]
    \item A fraction of the worst nests are abandoned and new nests are generated.
    \item Rank the solutions, find the global best and repeat.
\end{itemize}

\item Flower Pollination (FP): The FP \cite{FP} algorithm is based on the pollen movement of flowers. Agents (or pollen particles) move across the search space according to either a local or a global movement process, primarily centered around Lévy flight processes. A switch probability of $p$ determines which type of movement a particle will move by at various intervals.
\newline
\[
\begin{split}
    x_i^{t+1} &= x_i^t + \gamma L(\lambda)(g_* - x_i^t)\\
    x_i^{t+1} &= x_i^t + \varepsilon(x_j^t - x_k^t)\\
    L &\sim \frac{\lambda \tau(\lambda)\sin{(\pi\lambda/2)}}{\pi}\frac{1}{s^{1 + \lambda}}, \quad (s \gg s_0 > 0)   
\end{split}
\]

\end{itemize}

The MMO procedure runs all the above optimization algorithms concurrently and, at the end of every $x$ iterations, the procedure determines which optimizer agent is the current best (or has the current best solution for the problem at hand) out of the entire set/team and then updates the global current best solution for the whole team. This higher-level update will affect all of the algorithms that compose MMO, i.e., DE, PSO, PSOLévy, and BAT, BATLévy altering all the global best solution each is aware of with the team's best solution. The CS and FP algorithm agents are also affected by this communication process, but since they operate a little bit differently than the others, we must treat them with a bit of care. For both CS and FP, we choose an agent within their own respective internal population, at random (uniformly), and replace it with the current MMO best (this is done separately for each procedure). 

This communication process continues throughout the optimization process until a total number of generations is reached (or a convergence criterion is met). At the end point, MMO computes the best performing agent (algorithm) among the team and returns that particular agent's solution as the output of MMO.

\subsection{Communication Scheme} 
\label{subsec:scheme}
One of the most important features of MMO is its communication scheme. In this paper, we experiment with five different communication schemes that the master could use. These come in the form of higher-level operators applied to the global bests of each constituent algorithm agent. These operators include: averaging, weighted ranking (Rank Weighted), exponential weighted ranking (Exponential Weights), best ranking (Best Rank), and a meta-weighted communication scheme (Meta Weighted).

\begin{itemize}[wide, nosep, labelindent = 0pt, topsep = 1ex]
    \item Averaging: In this communication scheme, we compute the team's best aggregate solution by first recording the best performing solution found within each algorithm's own individual population, i.e., we extract each algorithm's global best $\mathbf{g}^*$, and the compute the mean of these individual global bests. This averaging is done so that incorporate the information fairly from each of the constituent optimization algorithms. However, this scheme could dilute the power of the parameter values but might lead to better exploration in subsequent generations.
    \item Rank Weighted: In this scheme we employ a weighted average of the algorithm team's global best values instead of just a simple averaging. To compute this weighted average, the agents' best solutions are ranked according to their fitness scores and assigned a weight according to their rank/placement in the list (the highest rank gets the largest weight, and so on and so forth).
    \item Exponential Weighted: This is a more restricted weighing scheme than rank weighting where we use exponential weights in order to compute the final team best solution. In this scheme, like in rank weighting, we sort the agent (best) solution vectors based on their fitness score and then apply an exponential weighting scheme to compute the individual (agent) weights. When working in the exponential scale, the best agent is assigned a much higher weight than the worst performing agents, i.e in our case given a rank vector $W^{*} = [7, 6, 5, 4, 3, 2, 1]$ exponential weights are computed as  , and $W^{e}_i = W^{*}[i]*\alpha^{W^{*}[1] - W^{*}[i]}$, where $W^{*}[1]=7$ and $\alpha=0.2$.
    \item Best Rank: As the name suggests, in this scheme we essentially just determine which agent algorithm's global solution is the best out of the team and throw out the rest. This is the most restrictive form of communication since we only keep track of the best agent solution (the ``best of the best'') found after $x$ iterations, and simply ignore all other information.
    \item Meta Weighted: This scheme computes the team best solution by averaging the solution value vectors returned by all other communication schemes described above in order to compute a single, final value (vector). 
\end{itemize}

\subsection{Communication Frequency}
\label{subsec:freq}
Once the communication scheme has been decided, we need to decide how often the master aggregator should communicate with the individual agent optimization algorithms within the team. This frequency is a hyper-parameter that the user should set, and could vary from forcing communication at the end of every single iteration or at the end of every $x$ iterations such that $x\leq Number \, of \, Generations$. From our preliminary experimentation, we found that the ideal range of communication frequency should be somewhere in the range $1 \leq x \leq 500$. The pseudocode for MMO algorithm is provided in Algorithm \ref{alg:mmo}.

\begin{algorithm}
\caption{Multi-Metaheuristic Optimization (MMO)}
\label{alg:mmo}
\begin{algorithmic}[1]
\State \textbf{Input: } Communication Frequency $\gamma$
\State Initialize $K$ individual optimization algorithm ``agents'', each with its own population size of $n$ randomly generated solution vectors 
\For{$x \gets 1$ to $NumGenerations$}  
    \State Run $K$ individual optimizers (1 simulation step)
    \If{$x \bmod \gamma = 0 $ }
        \State $\mathbf{g}^*_{team}$ = applyCommunicationScheme(getGlobalBests(team of $K$ Optimizers))
        \State For each optimizer $k$ in team, initialize its $\mathbf{g}^* = \mathbf{g}^*_{team}$
        \EndIf
        \EndFor
\State Select best solution $\mathbf{g}^*_{team}$ from $K$ individual optimizers
\end{algorithmic}
\end{algorithm}

\section{Experiments}
\label{sec:exp}
On three benchmark functions, we measure the individual performance of the various optimization algorithms that make up the agent team that our proposed Multi-Metaheuristic Optimizer(MMO) manages. We compare these measurements with MMO's performance under different hyper-parameter configurations. The three objective functions used are:

\begin{itemize}[wide, nosep, labelindent = 0pt, topsep = 1ex]

\item Rosenbrock Function: A non-convex function with a global minima of $0$ at $x = (1, 1, \ldots, 1)$. This function is defined as:  $f({x}) = \sum_{i = 1}^{D-1}[{100(x_{i+1} - x_i^2)^2 + (x_i - 1)^2}]$

\item Greiwank Function: A convex function with multiple local minima and a global minima of $0$ at $x = (0, 0, \ldots, 0)$. The function is defined as: $f(x) = \sum_{i=1}^{D}{\frac{x_i^2}{4000}} - \prod_{i =  1}^{D}{\cos{\left(\frac{x_i}{\sqrt{i}}\right)}} + 1$.

\item Zakharov Function: A plate shaped function with no local minima but a global minima of $0$ at $x = (0, 0, \ldots, 0)$. It is defined as $f(x) = \sum_{i=1}^{D}{x_i^2} + \left(\sum_{i=1}^{D}{0.5ix_i}\right)^2 + \left(\sum_{i=1}^{D}{0.5ix_i}\right)^4$

\end{itemize}

\subsection{Compositional vs Multi-Metaheuristic Optimizer}
First, we will analyze the performance of the individual optimization algorithms that compose MMO against MMO itself on a $15$-dimensional input space for the objective functions discussed above. Table \ref{results:single_v_mmo} shows the mean error and the standard error values returned by each single optimizer after iterating over $2000$ simulation steps using either $20, 50$, and $100$ agents. We observe that the individual algorithms do not do a good job at finding the global minima of the 15-dimensional Rosenbrock function which is a nonconvex function, with only CS, FP, and BATLévy appearing to converge to the global minimum. However, for the $15$-dimensional Griewank and Zakharov functions, all optimization algorithms appear to converge better, with CS and BATLévy appearing to be close to reaching the global minima. In most cases, the optimization algorithms perform better when the number of agents is increased.
\begin{table}[!htbp]
\centering
\caption{Individual algorithm performance versus MMO.}
\label{results:single_v_mmo}
\scalebox{0.6}{
\begin{tabular}{*8c}
\toprule
Objective function & Optimzer &  \multicolumn{2}{c}{20 agents} & \multicolumn{2}{c}{50 agents} & \multicolumn{2}{c}{100 agents}\\
\midrule
{} & {}   & mean error   & error se    & mean error   & error se & mean error   & error se\\
15D Rosenbrock & PSO   &  73282894 & 1954779 & 44243776&1552182 & 39063308  & 2239107\\
15D Rosenbrock & PSOLévy   &  71202029 & 2229271   & 68801471  & 1995423 & 35067503  & 892641\\
15D Rosenbrock & DE   &  108891 & 8084   & 11806  &  1740 & 20098  & 2584\\
15D Rosenbrock & BAT   & 1111747  &  140551  & 159600  & 19238 & 130034 & 19283\\
15D Rosenbrock & BATLévy   &  16.994  &  2.59   & 7.167  & 0.08 & 5.634  & 0.05\\
15D Rosenbrock & CS   &  48.40  &  2.247   & 26.024  & 1.538 & 11.051  & 0.411\\
15D Rosenbrock & FP  &  7.455  &  0.204   & 7.628  & 0.212 & 6.886  & 0.211\\
15D Griewank & PSO   &  7.108 & 0.129 & 7.218 & 0.066 & 5.915  & 0.089\\
15D Griewank & PSOLévy   &  7.407 & 0.116   & 7.126  & 0.071 & 6.689  & 0.066\\
15D Griewank & DE   &  0.967 & 0.01   & 0.902  &  0.006 & 0.921  & 0.004\\
15D Griewank & BAT   &  1.556  &  0.064  & 1.238  & 0.026 & 1.194 & 0.013\\
15D Griewank & BATLévy   &  0.319  & 0.015   & 0.173  & 0.017 & 0.162  & 0.013\\
15D Griewank & CS   &  0.032  &  0.001   & 0.016  & 0.001 & 0.01  & 0\\
15D Griewank & FP  &  0.007  &  0.001   & 0.003  & 0 & 0  & 0\\
15D Zakharov & PSO   &  927.06 & 212.2 & 136.4 & 4.93 & 99.35  & 1.602\\
15D Zakharov & PSOLévy   &  361.247 & 43.90 & 115.19  & 2.70 & 104.65  & 3.02\\
15D Zakharov & DE   & 7.31  & 0.44   & 24.718  &  0.797 & 35.174  & 0.784\\
15D Zakharov & BAT   &  203.85  &  12.7  & 171.75 & 11.473 & 134.81  & 24.24\\
15D Zakharov & BATLévy   &  0.001  & 0   & 0  & 0 & 0  & 0\\
15D Zakharov & CS   &  1.533  &  0.04   & 0.787  & 0.016 & 0.459  & 0.007\\
15D Zakharov & FP  &  0.006  &  0.001   & 0.07  & 0.003 & 0.1  & 0.005\\
\bottomrule
\end{tabular}}
\end{table}

In order to maintain consistency while using different optimization algorithms concurrently, MMO initializes all its constituent optimization algorithms with $100$ agents, this ensures all of them contribute equally to the optimization problem.  We tested MMO with different communication schemes and frequencies over $2000$ generations of iteration. We only include the Rank, Exponential, and Best communication schemes in Table \ref{results:mmo_analysis}, as the other two schemes did not perform as well as the others. Communication frequency was chosen from the set: $Frequency=\{1,10,50,500,1000,2000\}$. For the 15-dimensional Rosenbrock function, we observe that MMO outperforms its constituent algorithms by finding the global minimum when communication frequency is $1$ and communication scheme is either Exponential or Best Rank. For the 15-dimensional Griewank and Zakharov functions, MMO finds the global minimum in all cases. This indicates that MMO significantly outperforms the individual optimization algorithms. 
\begin{table}[!htbp]
\centering
\caption{Multi Metaheuristic Optimizer}
\label{results:mmo_analysis}
\scalebox{0.6}{
\begin{tabular}{*8c}
\toprule
Objective Function & Frequency &  \multicolumn{2}{c}{rank} & \multicolumn{2}{c}{Exponential} & \multicolumn{2}{c}{best}\\
\midrule
{} & {}   & mean error   & error se    & mean error   & error se & mean error   & error se\\
15D Rosenbrock & 1   &  7.09 & 0.732 &  0 & 0 &  0 & 0\\
15D Rosenbrock & 10   &  4.551 & 0.256   & 3.19  & 0.141 & 3.904  & 0.269\\
15D Rosenbrock & 50   &  4.757 & 0.131   & 4.801  &  0.236 & 3.904  & 0.269\\
15D Rosenbrock & 500   &  5.596  &  0.099   & 5.061 & 0.083 & 4.066  & 0.182\\
15D Rosenbrock & 1000   & 5.051  & 0.173 & 5.1471  & 0.058 & 4.784  & 0.255\\
15D Rosenbrock & 2000   &  5.469  & 0.136 & 5.783  &  0.215  & 4.36  & 0.268\\
15D Griewank & 1   &  0.01 & 0.001 &  0.016 & 0.001 &  0.009 & 0.001\\
15D Griewank & 10   &  0.001 & 0   & 0.004  & 0.001 & 0.006  & 0.001\\
15D Griewank & 50   &  0 & 0   & 0.001  &  0 & 0  & 0\\
15D Griewank & 500   &  0  &  0   & 0 & 0 & 0.002  & 0\\
15D Griewank & 1000   &  0  &  0   & 0 & 0 & 0.002  & 0\\
15D Griewank & 2000   &  0  &  0   & 0.001 & 0 & 0.002  & 0\\
15D Zakharov & 1   &  0.001 & 0 &  0 & 0 &  0 & 0\\
15D Zakharov & 10   &  0 & 0   & 0  & 0 & 0  & 0\\
15D Zakharov & 50   &  0 & 0   & 0  & 0 & 0  & 0\\
15D Zakharov & 500   &  0 & 0   & 0  & 0 & 0  & 0\\
15D Zakharov & 1000   &  0 & 0   & 0  & 0 & 0  & 0\\
15D Zakharov & 2000   &  0 & 0   & 0  & 0 & 0  & 0\\
\bottomrule
\end{tabular}}
\end{table}

\subsection{Ablation Test}
Since MMO contains multiple optimization algorithms running concurrently, we conduct an ablation study (Table \ref{results:ablation}) to understand the contribution of each individual optimization algorithms to MMO's overall performance and to gain a deeper understanding of MMO's overall functioning. We conduct this ablation test by removing each individual optimization algorithm from it at a time and then subsequently optimizing the partially-reduced MMO algorithm on the 15D Rosenbrock objective function, with communication frequency of $1$ and an 'Exponential' communication scheme for $2000$ generations. Based on the table below DE, PSO, BAT and CS contribute the least to MMO's performance while the FP algorithm contributes the most. According to Table \ref{results:single_v_mmo}, we observe the best performing optimization algorithm was BATLévy, however the most important metaheuristic for MMO, based on the ablation study is flower pollination, FP. This shows that different algorithms bring diversified information into the optimization process, which is crucial for faster convergence. In sum, the architecture of the MMO algorithm ensures that it performs much better than the sum of its parts, even if some parts contribute more than others. 
\begin{table}[!htbp]
\centering
\caption{15D Rosenbrock Ablation}
\label{results:ablation}
\scalebox{0.7}{
\begin{tabular}{*3c}
\toprule
Ablation & mean error & error se \\
\midrule
PSO   &  0.40 & 0.125 \\
PSOLévy   &  0.802 & 0.1683 \\
DE   &  0.0186 & 0.001 \\
BAT   &  0.040 & 0.005 \\
BATLévy   &  0.96 & 0.131 \\
CS   &  0.068 & 0.007 \\
FP  &  5.52 & 0.08 \\
No Ablation & 0 & 0\\
\bottomrule
\end{tabular}}
\end{table}

\subsection{Cross Dimension Test}
MMO tends to outperform other optimization algorithms for higher dimensional problems. To test this property we compared the performance of MMO against BATLévy on 5, 10, 15 and 25D Rosenbrock functions. We choose BATLévy for comparison since it was the best performing individual optimization algorithm on the 15D Rosenbrock function in the experiment earlier. In the table below we observe that BATLévy appears to converge for lower dimensional Rosenbrock functions, however, when the dimensionality increases, it struggles to find the global minimum. The MMO algorithm, on the other hand, has no trouble converging to the global minimum even for the 25D Rosenbrock function. This shows that the performance and value of the MMO becomes more apparent for higher dimensional problems when conventional, individual algorithms struggle. 
\begin{table}[!htbp]
\centering
\caption{Performance versus increasing dimensions of the Rosenbrock function. }
\scalebox{0.6}{
\begin{tabular}{*6c}
\toprule
Dimensions & Num Generations &\multicolumn{2}{c}{BATLévy} & \multicolumn{2}{c}{MMO}\\
\midrule
{} & {} & mean error   & error se & mean error   & error se\\
5D-Rosenbrock & 2000 &  0.005 & 0.001   & 0  & 0\\
10D-Rosenbrock & 2000 &  1.155 & 0.227   & 0  & 0\\
15D-Rosenbrock & 2000 &  5.634  &  0.051   & 0  & 0\\
25D-Rosenbrock & 4000 &  27.332  &  3.131   & 0.098  & 0.005\\
\bottomrule
\end{tabular}}
\end{table}

\subsection{SVM with SGD and MMO}
To test a real-world application of the MMO algorithm, we compare the performance of MMO with stochastic gradient descent (SGD) to optimize the parameters of a (linear kernel) of a support vector machine (SVM). We test both the algorithms on two data sets taken from the UCI Machine Learning Repository, 

\begin{itemize}[wide, nosep, labelindent = 0pt, topsep = 1ex]
\item Breast Cancer Wisconsin Original Data Set (BCW): This data set contains 699 instances of data, 10 input attributes and 2 output classes i.e. benign and malignant. 
\item Image Segmentation Data Set  (IS): This data set contains 2310 instances of data, 19 input attributes and 7 output classes i.e. brickface, sky, foliage, cement, window, path, and grass
\end{itemize}

\begin{table}[!htbp]
\centering
\caption{SVM SGD vs MMO Hyper-Parameter Optimization}
\label{results:svm}
\scalebox{0.6}{
\begin{tabular}{*9c}
\toprule
Dataset & \multicolumn{2}{c}{config} &  \multicolumn{3}{c}{SGD} & \multicolumn{3}{c}{MMO}\\
\midrule
{} & SGD Config & MMO Config  & loss & train acc & valid acc  & loss & train acc & valid acc\\
BCW & $\lambda=0,\alpha=0.01$ & $\lambda=0,f=1,best$  &  0.366 & 88.44\% & 86.71\%  & 0.334 & 88.26\% & 89.28\%\\
BCW & $\lambda=0,\alpha=0.001$ & $\lambda=0,f=100,best$  &  0.377 & 87.75\% & 85.64\%  & 0.334 & 88.22\% & 89.16\%\\
BCW & $\lambda=0.01,\alpha=0.01$ & $\lambda=0,f=1,best$  &  0.3813 & 88.21\% & 86.21\%  & 0.334 & 88.18\% & 89.04\%\\
IS & $\lambda=0.1,\alpha=0.001$  & $\lambda=0,f=1,exponential$  &  1.78 & 73.67\% & 71.73\%  & 0.8142 & 89.79\% & 86.74\%\\
IS & $\lambda=0.01,\alpha=0.001$  & $\lambda=0,f=10,Exponential$  &  1.88 & 75.69\% & 74.20\%  & 1.388 & 88.96\% & 90.76\\
IS & $\lambda=0.001,\alpha=0.001$  & $\lambda=0,f=1,best$  &  1.99 & 73.98\% & 70.99\%  & 1.50 & 88.07\% & 87.12\%\\
\bottomrule
\end{tabular}}
\end{table}

For each data set the train-validation-test split is 60-20-20. In Table \ref{results:svm} above, we show the 3 best hyper-parameter configurations obtained by SGD and MMO on Breast Cancer Wisconsin and Image Segmentation data sets, where $\alpha$ is the learning rate, $\lambda$ is the regularization rate, and $f$ is frequency of communication. Both optimizers ran for 1000 iterations to maintain consistency of results for comparison. 
Based on the table above we see that MMO outperforms SGD on both Breast Cancer Wisconsin and Image Segmentation datasets, in-terms of loss and validation accuracy scores. Using the best hyper-parameter configurations we analyze the performance of both algorithms on the test datasets. The plots below show the trajectory of the loss function with respect to the number of iterations. We observe that for both the data sets (BCW and IS) MMO obtains minimum loss in the matter of a few iterations, whereas SGD continues to minimize the loss even after 1000 iterations. The trajectory of the MMO is a lot smoother than SGD, with little to no variance, which is not the case with SGD whose loss trajectories are very volatile. This indicates that MMO is more robust than SGD and is less likely to overfit during the training process.

\begin{figure}[!htb]
\minipage{0.5\textwidth}
\includegraphics[scale=0.4]{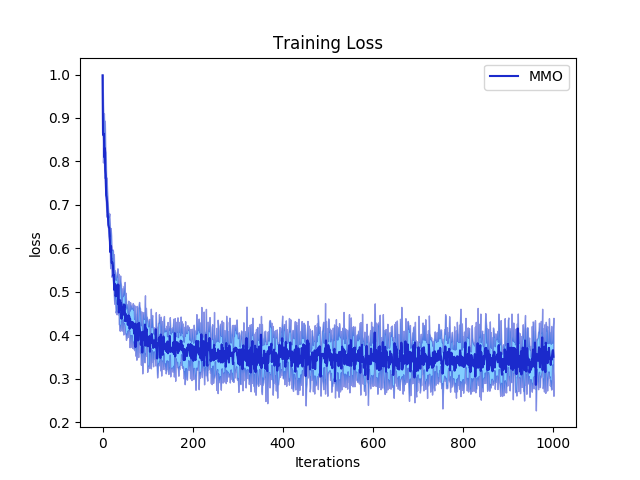}
\caption{BCW SGD Loss}\label{fig:sgd_loss_wisc}
\endminipage\hfill
\minipage{0.5\textwidth}
\includegraphics[scale=0.4]{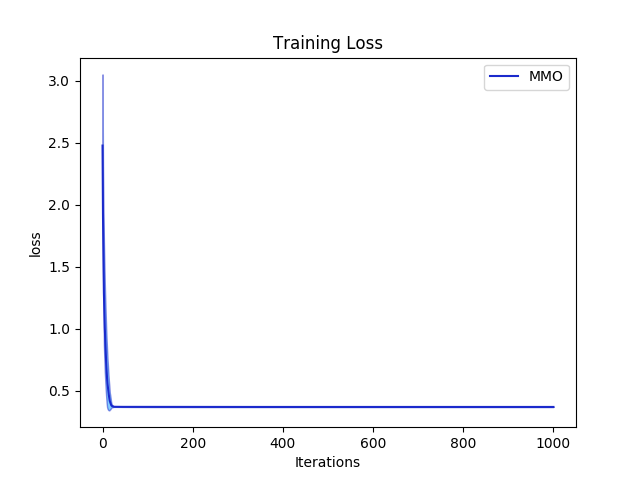}
\caption{BCW MMO Loss}\label{fig:mmo_loss_wisc}
\endminipage\hfill
\end{figure}
\clearpage
\begin{figure}[!htb]
\minipage{0.5\textwidth}
\includegraphics[scale=0.4]{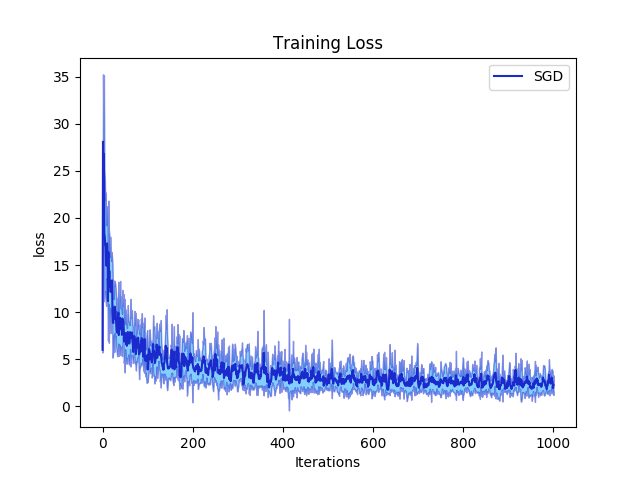}
\caption{IS SGD Loss}\label{fig:sgd_loss_is}
\endminipage\hfill
\minipage{0.5\textwidth}
\includegraphics[scale=0.4]{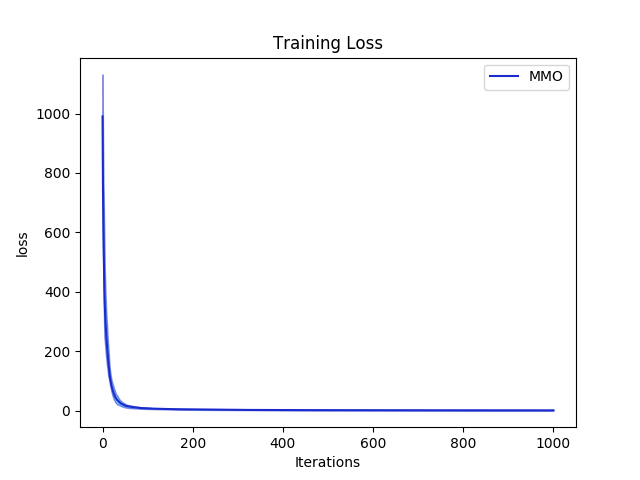}
\caption{IS MMO Loss}\label{fig:mmo_loss_is}
\endminipage\hfill
\end{figure}
 The final train and test accuracy results for both data sets are shown in the table below,
 \begin{table}[!htbp]
\centering
\caption{SVM SGD vs MMO Train and Test performance}
\scalebox{0.7}{
\begin{tabular}{*9c}
\toprule
Dataset & \multicolumn{2}{c}{config} &  \multicolumn{3}{c}{SGD} & \multicolumn{3}{c}{MMO}\\
\midrule
{} & SGD Config & MMO Config  & loss & train acc & test acc  & loss & train acc & test acc\\
BCW & $\lambda=0,\alpha=0.01$ & $\lambda=0,f=1,best$  &  0.35 & 87.76\% & 84.07\%  & 0.37 & 87.11\% & 89.14\%\\
IS & $\lambda=0.1,\alpha=0.001$  & $\lambda=0,f=1,exponential$  &  2.24 & 72.85\% & 71.99\%  & 0.953 & 91.14\% & 91.24\%\\
\bottomrule
\end{tabular}}
\end{table}
\section{Conclusion}
\label{sec:conclusion}
Our proposed Multi-Metaheuristic Optimizer (MMO) provides an effective way of combining different population-based metaheuristics, outperforming other optimization algorithms by a clear margin in terms of speed and accuracy. Due to MMO's master/slave configuration, multiple slaves can be concurrently implemented to speed up the optimization process. As multiple optimization algorithms are involved in the decision-making process at any given time, MMO is more likely to provide robust results. It is flexible given that the approach towards optimization can be tweaked using two parameters i.e. by selecting from a plethora of communication schemes as well as choosing the frequency of communication (i.e high frequency: favors exploitation while low frequency: favors exploration). 
Due to the flexible architecture of MMO, one can further finetune the algorithm by dynamically altering the communication process, i.e. increase communication frequency when the optimizer stops learning and decrease or fix the frequency otherwise. 

\end{document}